\documentclass{article}
\usepackage{spconf,amsmath,epsfig}
\usepackage{graphicx} 
\usepackage{booktabs}
\usepackage{hyperref}
\usepackage{amsmath,amsfonts}
\usepackage{subfigure,multirow}
\usepackage{arydshln}
\let\OLDthebibliography\thebibliography
\renewcommand\thebibliography[1]{
  \OLDthebibliography{#1}
  \setlength{\parskip}{0pt}
  \setlength{\itemsep}{0pt plus 0.3ex}
}

\hypersetup{
    colorlinks=true,
    linkcolor=blue,
    filecolor=magenta,      
    urlcolor=magenta,
    citecolor=cyan,
    pdftitle={Overleaf Example},
    pdfpagemode=FullScreen,
    }

\begin{document}\sloppy

\def\x{{\mathbf x}}
\def\L{{\cal L}}

\title{Q-Boost: On Visual Quality Assessment Ability of Low-level Multi-Modality Foundation Models}
%

\names{Zicheng Zhang$^{1*}$ \quad Haoning Wu$^{2*}$  \quad Zhongpeng Ji$^3$  \quad Chunyi Li$^1$ \quad Erli Zhang$^2$ \quad Wei Sun$^1$}{{Xiaohong Liu$^1$ \quad Xiongkuo Min$^1$ \quad Fengyu Sun$^3$ \quad Shangling Jui$^3$ \quad Weisi Lin$^2$ \quad Guangtao Zhai$^1$}}{{\url{https://github.com/Q-Future/Q-Instruct/boost_qa}}}
\address{$^1$Shanghai Jiaotong University, $^2$S-Lab, Nanyang Technological University, $^3$Huawei}

\maketitle

%
\begin{abstract}
Recent advancements in Multi-modality Large Language Models (MLLMs) have demonstrated remarkable capabilities in complex high-level vision tasks. However, the exploration of MLLM potential in visual quality assessment, a vital aspect of low-level vision, remains limited. To address this gap, we introduce \textbf{Q-Boost}, a novel strategy designed to enhance low-level MLLMs in image quality assessment (IQA) and video quality assessment (VQA) tasks, which is structured around two pivotal components: 1) \textbf{Triadic-Tone Integration}: Ordinary prompt design simply oscillates between the binary extremes of $positive$ and $negative$. \textbf{Q-Boost} innovates by incorporating a `middle ground' approach through $neutral$ prompts, allowing for a more balanced and detailed assessment. 2) \textbf{Multi-Prompt Ensemble}:  Multiple quality-centric prompts are used to mitigate bias and acquire more accurate evaluation. The experimental results show that the low-level MLLMs exhibit outstanding zeros-shot performance on the IQA/VQA tasks equipped with \textbf{Q-Boost} strategy. 
\end{abstract}

\section{Introduction}
\label{sec:intro}
Multi-modality Large Language Models (MLLMs) have exhibited strong capabilities in general visual perception and understanding, which have been tested and confirmed in various vision-language tasks, including image captioning, visual question answering, cross-modality grounding, as well as standard vision tasks like image classification or segmentation. While the focus has largely been on high-level visual content perception and understanding, the proficiency of MLLMs in visual quality assessment remains less understood. 

\begin{figure}
    \centering
    \includegraphics[width=\linewidth]{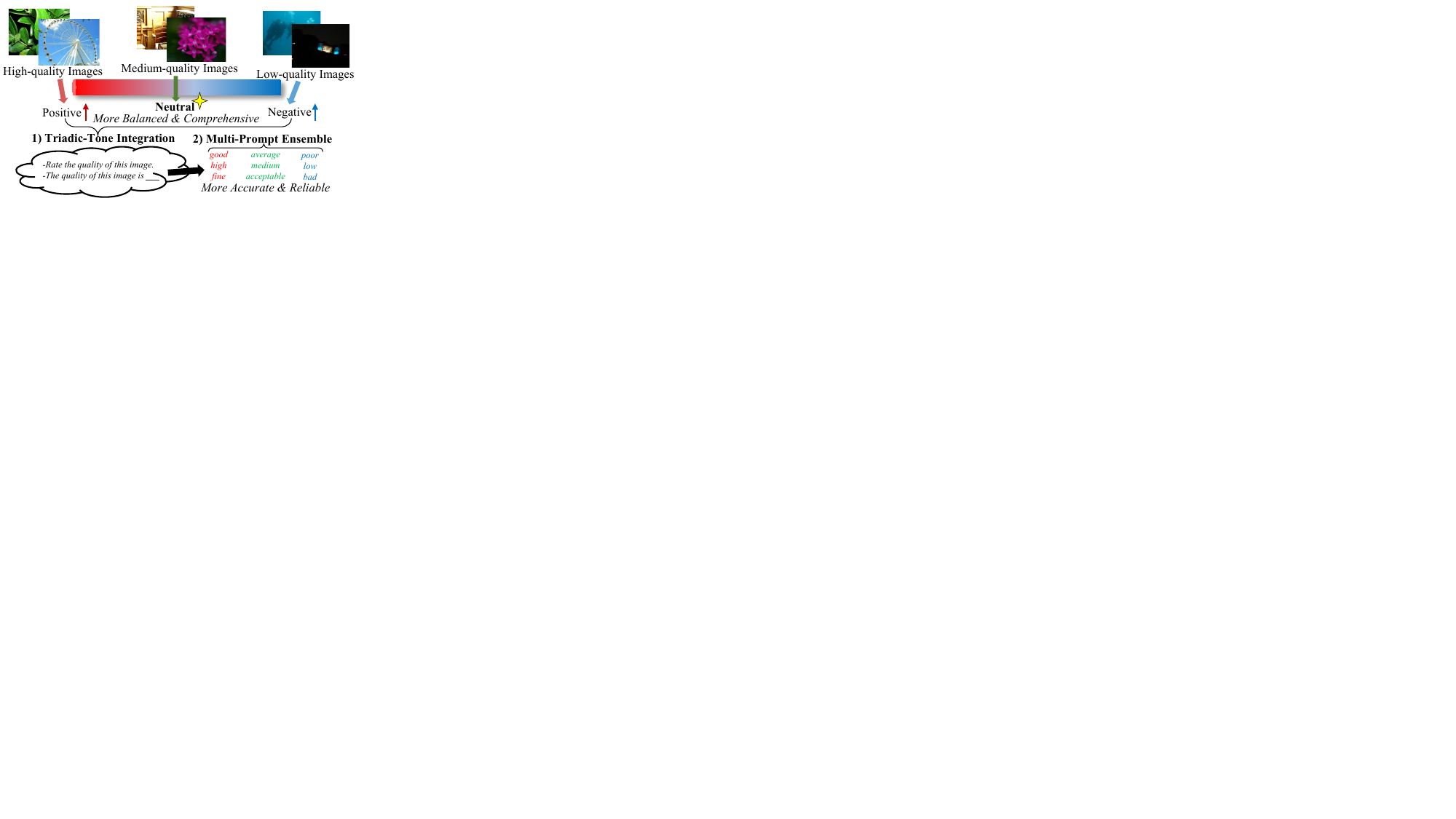}
    \caption{Motivation of \textbf{Q-Boost}. The \textbf{Triadic-Tone Integration} strategy helps provide a more balanced and comprehensive assessment while the \textbf{Multi-Prompt Ensemble} strategy helps improve the accuracy and reliability of evaluation. }
    \label{fig:motivation}
\end{figure}

Over the past decade, a significant amount of research has been dedicated to exploring image quality assessment (IQA) and video quality assessment (VQA). In the initial phases, researchers extract handcrafted, quality-aware features from images or videos, subsequently regressing these features to derive quality scores~\cite{mittal2012making,zhang2015feature,zhang2021ano}. Subsequently, the advent and success of deep neural networks (DNNs) have paved the way for the use of deep-learning models to extract and regress quality features into corresponding scores~\cite{dbcnn,su2020blindly,ke2021musiq,vsfa,wu2023dover,Zhang_2023_CVPR}. Regardless of the extraction technique, these quality assessment methodologies predominantly depend on mean opinion scores (MOSs). The fundamental learning approach involves minimizing the discrepancy between these features and MOSs to obtain insightful knowledge for quality assessment. While MOS regression is straightforward and enables models to directly assess quality, it suffers from a lack of interpretability. In contrast, the text-supervised MLLM can output quality assessment results in natural language format and make explanations, which operate more similarly as human perception. Given the rapid advancements in MLLM, integrating it into quality assessment tasks is not only meaningful but also expands the horizon beyond traditional MOS regression. This approach facilitates training AI through text supervision to gain quality knowledge and designing prompts for MLLM to predict quality scores effectively.

\begin{figure*}
    \centering
    \includegraphics[width=.84\linewidth]{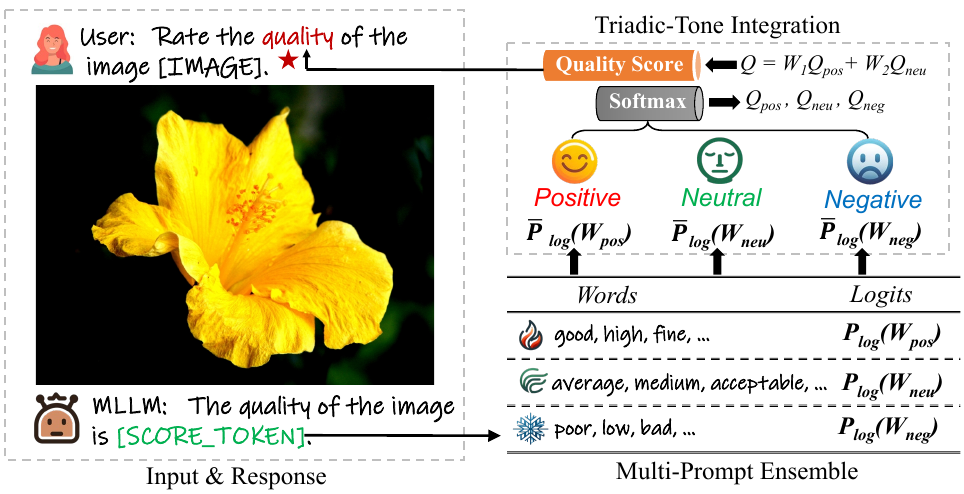}
    \caption{The framework of the \textbf{Q-Boost}. The image and prompt are fed into the MLLM, where the log probabilities (\textit{logits}) are computed between the \textit{[SCORE\_TOKEN]} and triadic-tone words (including ensembles of multiple prompts). Then the \textit{logits} of different tones are put through softmax operation and integrated into the zero-shot quality score with weighted average pooling.}
    \label{fig:framework}
    \vspace{-0.3cm}
\end{figure*}

The design of prompts like ``\textit{Rate the quality of the image}'' and ``\textit{Score the quality of the image from 1 to 5}'' is a natural approach for eliciting feedback from MLLMs regarding image quality. Nevertheless, these strategies have been shown to exhibit bias~\cite{wu2023qbench}, as responses from MLLMs are often positively skewed and extreme. It is more recommended to transfer the IQA task into a binary classification problem on the probabilities of responding ``\textit{Rate the quality of the image}'' with \textit{good} and \textit{bad} from the MLLMs, where the probabilities are quantified into quality scores via softmax. 
This softmax-based evaluation strategy has been used to benchmark the assessment abilities of various MLLMs~\cite{wu2023qbench} and even revealed that MLLMs are actually effective zero-shot assessment assistants after multi-modality low-level dataset tuning~\cite{wu2023qinstruct}.

To further enhance this methodology, we introduce \textbf{Q-Boost}, a novel advancement designed to significantly improve the performance of MLLMs on IQA~\cite{ke2021musiq,dbcnn,wang2023exploring} as well as IQA-based video quality assessment (VQA)~\cite{li2019vsfa,wu2023maxvqa,wu2023fastervqa,wu2022fastvqa,wu2023discovqa}. \textbf{Q-Boost} consists of two strategies: 1) The traditional prompt design, which typically fluctuates between the binary extremes of $positive$ and $negative$, neglects the fact that some images might be of medium quality. \textbf{Q-Boost} proposes a significant innovation \textbf{Triadic-Tone Integration} (TTI) in this context, by introducing $neutral$ prompts such as \textit{average}, \textit{medium}, and \textit{acceptable}. This not only enables a more balanced and comprehensive assessment but also enriches the evaluation process with greater detail, moving beyond the simplistic binary paradigm. 2) \textbf{Q-Boost} presents \textbf{Multi-Prompt Ensemble} (MPE) to enhance the accuracy of the assessment results. On one hand, multi-prompt expands the semantic scope, capturing nuances that a single word may overlook. On the other hand, multi-prompt reduces ambiguity as different terms carry slightly varied connotations, leading to more reliable evaluation. The experimental results show that the MLLM with \textbf{Q-Boost} achieves state-of-the-art zero-shot performance on both IQA/VQA datasets.

\begin{table*}[!t]
\centering
\renewcommand\arraystretch{1.2}
\resizebox{\linewidth}{!}{\begin{tabular}{cccc:cccc:cccc:cccc}
\toprule
\multicolumn{4}{c:}{\includegraphics[width=7cm]{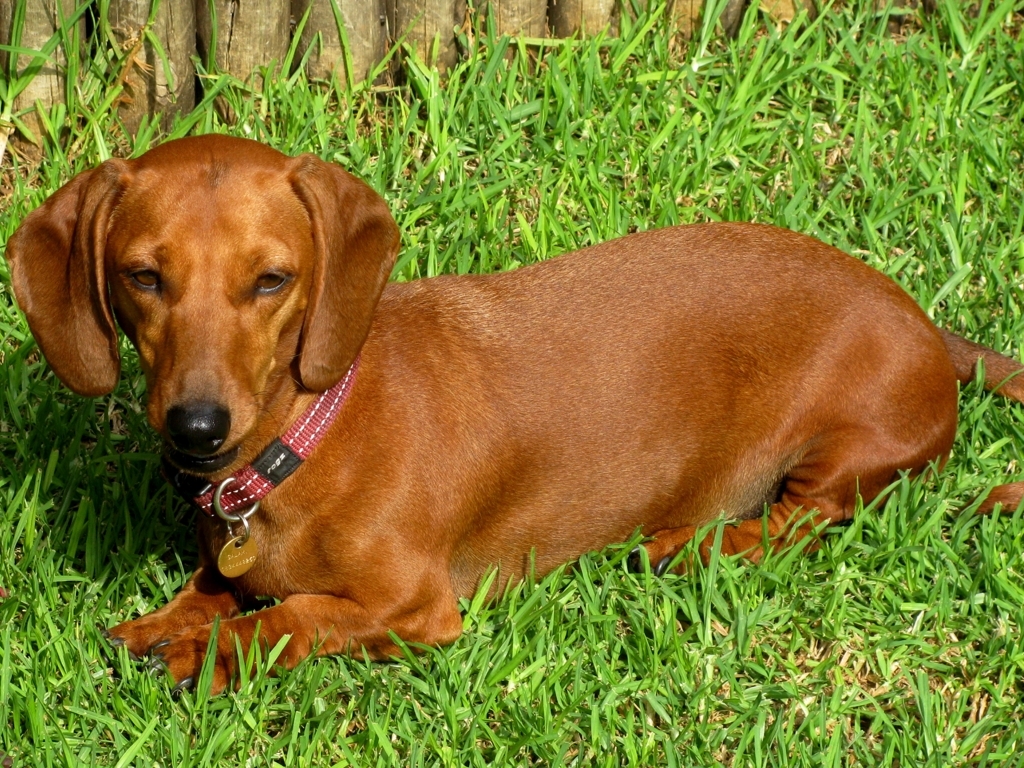}} & \multicolumn{4}{c:}{\includegraphics[width=7cm]{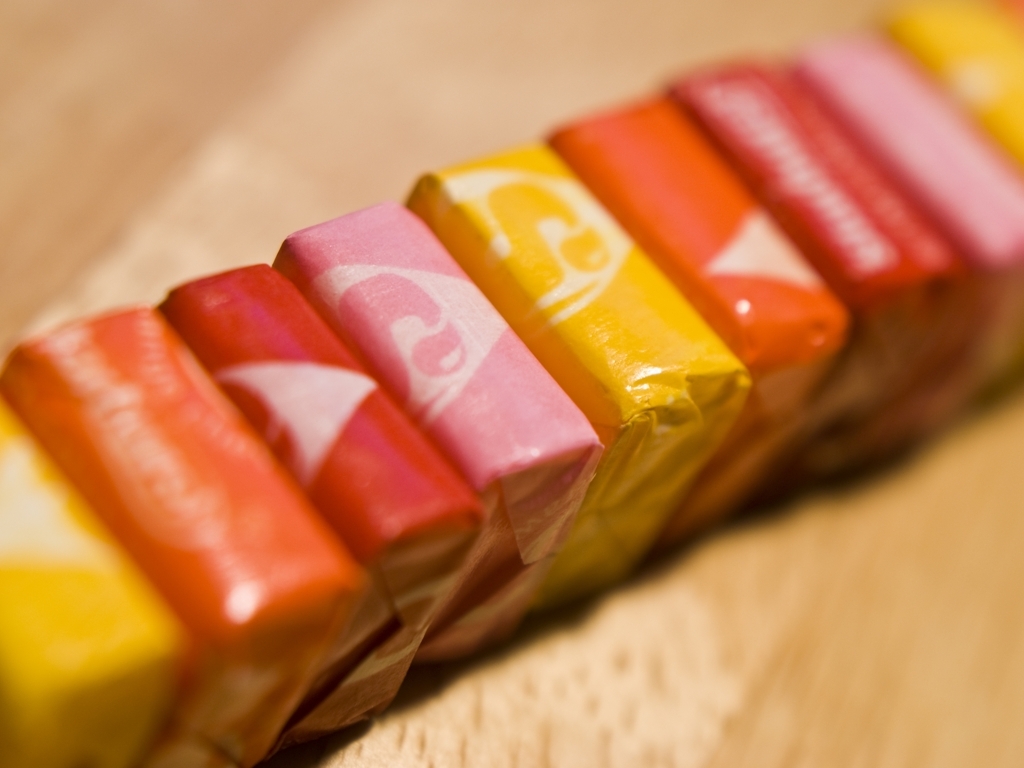}} & \multicolumn{4}{c:}{\includegraphics[width=7cm]{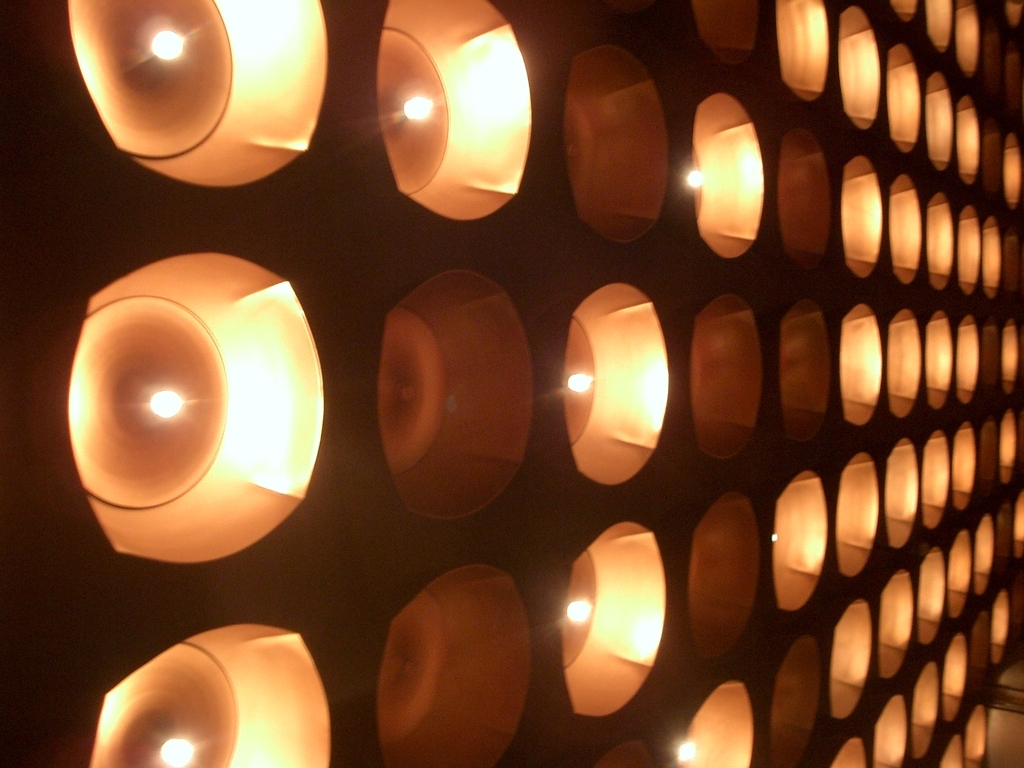}} & \multicolumn{4}{c}{\includegraphics[width=7cm]{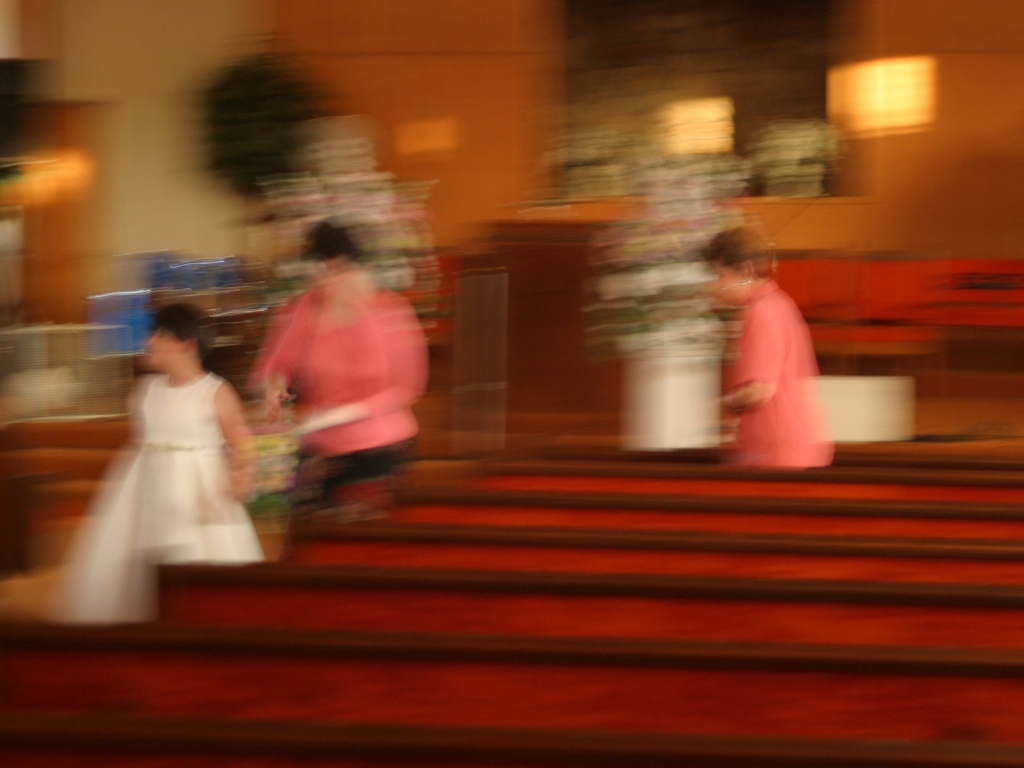}}\\ \hline
$\mathcal{{P}}_{log}(good)$ & $\mathcal{{P}}_{log}(average)$ & $\mathcal{{P}}_{log}(poor)$ & GT & $\mathcal{{P}}_{log}(good)$ & $\mathcal{{P}}_{log}(average)$ & $\mathcal{{P}}_{log}(poor)$ & GT & $\mathcal{{P}}_{log}(good)$ & $\mathcal{{P}}_{log}(average)$ & $\mathcal{{P}}_{log}(poor)$ & GT & $\mathcal{{P}}_{log}(good)$ & $\mathcal{{P}}_{log}(average)$ & $\mathcal{{P}}_{log}(poor)$ & GT\\ \hdashline
\textbf{11.14} & 8.68 & 7.96 & 4.12/5 & 10.02 & \textbf{10.71} & 9.07 & 3.11/5 & 8.92 & \textbf{10.75} & 9.59 & 2.51/5 & 6.46 & 8.62 & \textbf{10.70} & 1.10/5 \\
\bottomrule
\end{tabular}}
\caption{Example log probabilities (\textit{logits}) of \textit{good}, \textit{average}, and \textit{poor}. It is evident that images of lower quality exhibit a stronger correlation with \textit{negative} prompts, whereas those of higher quality demonstrate a greater alignment with \textit{positive} prompts. Images classified as being of medium quality tend to show increased \textit{neutral} \textit{logits}.}
\label{tab:example}
\end{table*}

\section{The Q-Boost}
The assessment framework of MLLM (\textbf{Q-Boost}) is illustrated in Fig. \ref{fig:framework}. 
First, we describe the process of getting log probabilities (\textit{logits})
of certain prompts from MLLM. The image along with the prompt ``\textit{Rate the quality of the image. The quality of this image is [SCORE\_TOKEN]}'' is input into the MLLM and the \textit{logits} can be derived as the probability distance between \textit{[SCORE\_TOKEN]} and the certain prompt words such \textit{good} and \textit{bad}:

\begin{equation}
    \mathcal{P}_{log}(\mathcal{W}) = \mathrm{MLLM}(\mathcal{T},\mathcal{I}),
\end{equation}
where $\mathcal{P}_{log}(\mathcal{W})$ represents the log probabilities of certain prompt words, $\mathrm{MLLM}(\cdot)$ indicates the MLLM assessment operation, $\mathcal{T}$ stands for the input prompt text, and $\mathcal{T}$ denotes the input image.

\subsection{Triadic-Tone Integration (TTI)}
\label{sec:integration}
In many text-prompted quality assessment works~\cite{wang2023exploring,wu2023bvqi}, only binary extremes (\textit{positive} and \textit{negative}) are taken into consideration, which is difficult to distinguish the quality of images with medium quality as shown in Table \ref{tab:example}.  Therefore, we introduce a third tone \textit{neutral} to such as \textit{average}, \textit{medium}, and \textit{acceptable} to provide a more balanced and comprehensive evaluation:
\begin{equation}
\begin{aligned}
    \mathcal{Q}_{pos}, & \mathcal{Q}_{neu}, \mathcal{Q}_{neg} = \\
    &\gamma(\mathcal{\bar{P}}_{log}(\mathcal{W}_{pos}),\mathcal{\bar{P}}_{log}(\mathcal{W}_{neu}),\mathcal{\bar{P}}_{log}(\mathcal{W}_{neg})), \\
\end{aligned}
\end{equation}
where $\mathcal{\bar{P}}_{log}(\mathcal{W}_{pos}),\mathcal{\bar{P}}_{log}(\mathcal{W}_{neu}),\mathcal{\bar{P}}_{log}(\mathcal{W}_{neg})$ represent the averaged \textit{logits} of \textit{positive}, \textit{neutral}, \textit{negative} prompt words (see in Sec.~\ref{sec:ensemble}), $\gamma(\cdot)$ indicates the softmax function. Then the integrated quality scores can be obtained as follows: 
\begin{equation}
    \mathcal{Q} = w_1 \mathcal{Q}_{pos} + w_2 \mathcal{Q}_{neu},
\end{equation}
where $w_1=1.0$ and $w_2=0.5$ are the weighted parameters for $\mathcal{Q}_{pos}$ and $\mathcal{Q}_{neu}$.

\subsection{Multi-Prompt Ensemble (MPE)}
\label{sec:ensemble}
Generally speaking, multi-prompt enhance the semantic range, grasping subtleties that a single word might miss. Conversely, they minimize ambiguity since different terms have slightly different implications, resulting in a more dependable evaluation.
Therefore, we propose to use a group of synonym words instead of a single word to describe one certain tune:
\begin{equation}
    \bar{\mathcal{P}}_{log}(\mathcal{W}_{\alpha})=\frac{1}{n_\alpha}\sum {\mathcal{P}}_{log}(\mathcal{W}_{\alpha}),
\end{equation}
where $\alpha \in \{pos,neu,neg\}$, $\bar{\mathcal{P}}_{log}(\mathcal{W}_{\alpha})$ represents the averaged \textit{logits} of tune $\alpha$, and $n_\alpha$ stands for the number of the words for describing tune $\alpha$.

\subsection{Video Evaluation}
The process described above can directly get the quality score of a single image. For videos, we extract the frames every second and average the frames' \textit{logits} as the video \textit{logits}:
\begin{equation}
    \bar{\mathcal{P}}_{log}^v(\mathcal{W}_{\alpha}) = \frac{1}{n_{v}} \sum_{i=1}^{n_{v}} \bar{\mathcal{P}}_{log}^i(\mathcal{W}_{\alpha}),
\end{equation}
where $\alpha \in \{pos,neu,neg\}$, $n_{v}$ indicates the length of the video, $\bar{\mathcal{P}}_{log}^i(\mathcal{W}_{\alpha})$ represents the $i$-th frame's \textit{logits}, and $\bar{\mathcal{P}}_{log}^v(\mathcal{W}_{\alpha})$ stands for the video's \textit{logits}. Then the video quality can be calculated with the same interaction process described above.

\begin{table}[t]
    \centering
    \renewcommand\arraystretch{1.1}
    \caption{Review of the IQA/VQA datasets.}
    \resizebox{\linewidth}{!}{\begin{tabular}{c|l|c|c}
    \toprule
        \textbf{Type} & \textbf{Dataset} & \textbf{Num.} & \textbf{Content} \\ \hline
        \multirow{3}{*}{IQA} & KonIQ-10k~\cite{koniq} & 10,741 & In-the-wild \\
        & KADID-10K~\cite{kadid} & 10,125 & Artificially Distorted \\
        & CGIQA-6K~\cite{zhang2023subjective} & 6,000 & Computer Graphics Images \\ \hdashline
        \multirow{3}{*}{VQA} & LIVE-VQC~\cite{vqc} & 585 & In-the-wild \\
        & KoNViD-1k~\cite{kv1k} & 1,200 & In-the-wild \\
        & Youtube-UGC~\cite{ytugccc} & 1,380 & In-the-wild \\
        & MaxWell (val)~\cite{wu2023maxvqa} & 909 & In-the-wild \\
    \bottomrule   
    \end{tabular}} 
    \label{tab:datasets}
\end{table}

\begin{table}[]
    \centering
    \renewcommand\arraystretch{1.1}
    \renewcommand\tabcolsep{3pt}
    \caption{Results on the IQA datasets. Both mPO-7B (\textbf{Q-Instruct}) and mPO-7B (\textbf{Q-Boost}) utilize the \textbf{Q-Instruct}~\cite{wu2023qinstruct}-tuned weights. mPO-7B (\textbf{Q-Instruct}) is validated without the \textbf{Q-Boost} strategy. About \textit{48.92\%} images of the KonIQ-10k are included in the \textbf{Q-Instruct} and \textit{0\%} for other datasets. Best zero-shot performance is marked in \bf{bold}.}
    \resizebox{\linewidth}{!}{\begin{tabular}{l|cc:cc:cc} \toprule
         \textbf{Datasets} & \multicolumn{2}{c:}{\textit{KonIQ-10k}} & \multicolumn{2}{c:}{\textit{KADID-10K}} & \multicolumn{2}{c}{\textit{CGIQA-6K}}\\ \hdashline
         \textbf{Model} & SRCC & PLCC & SRCC & PLCC & SRCC & PLCC \\ \hline
         \multicolumn{7}{l}{\textbf{Opinion-Aware Approaches}} \\ \hdashline
         DBCNN~\cite{dbcnn} & 0.884 & 0.875 & 0.856 & 0.851 & 0.662 & 0.664\\ 
         HyperIQA~\cite{su2020blindly} & 0.917 & 0.906 & 0.845 & 0.852 & 0.636 & 0.660\\ 
         MUSIQ~\cite{ke2021musiq} & 0.915 & 0.937 & 0.572 & 0.584 & 0.625 & 0.641\\ \hdashline
         \multicolumn{7}{l}{\textbf{Zero-shot Approaches}} \\ \hdashline
         NIQE~\cite{mittal2012making} & 0.316 & 0.377 & 0.374 & 0.428 & 0.075 & 0.056 \\
         IL-NIQE~\cite{zhang2015feature} & 0.537 & 0.523 & 0.558 & 0.534 & 0.108 & 0.082 \\
         CLIP-IQA~\cite{wang2023exploring} & 0.468 & 0.505 & 0.501 & 0.520 & 0.294 & 0.300 \\ \hdashline
          mPO-7B (\textbf{Q-Instruct}~\cite{wu2023qinstruct}) & 0.902 & 0.888 & 0.698 & 0.684 & 0.629 & 0.644 \\
          mPO-7B (\textbf{Q-Boost}) & \bf{0.912} & \bf{0.923} & \bf{0.702} & \bf{0.699} & \bf{0.630} & \bf{0.648}\\
          \bottomrule   
    \end{tabular}}
    \label{tab:iqa}
\end{table}

\begin{table*}[]
    \centering
    \renewcommand\arraystretch{1.1}
    \renewcommand\tabcolsep{10pt}
    \caption{Results on the VQA datasets. The best and second zero-shot performance are marked in \textbf{bold} and \underline{underline}.}
    \resizebox{\linewidth}{!}{\begin{tabular}{l|cc:cc:cc:cc} \toprule
         \textbf{Datasets} & \multicolumn{2}{c:}{\textit{LIVE-VQC}} & \multicolumn{2}{c:}{\textit{ KoNViD-1k}} & \multicolumn{2}{c:}{\textit{YouTube-UGC}} & \multicolumn{2}{c}{\textit{MaxWell (val)}}\\ \hdashline
         \textbf{Model} & SRCC & PLCC & SRCC & PLCC & SRCC & PLCC & SRCC & PLCC\\ \hline
         \multicolumn{7}{l}{\textbf{Opinion-Aware Approaches}} \\ \hdashline
         TLVQM~\cite{tlvqm} & 0.799 & 0.803 & 0.773 & 0.768 & 0.669 & 0.659 & 0.661 & 0.652\\
         VSFA~\cite{vsfa} & 0.773 & 0.795 & 0.773 & 0.775 & 0.724 & 0.743 & 0.676 & 0.678\\
         VIDEVAL~\cite{videval} & 0.752 & 0.751 & 0.783 & 0.780 & 0.779 & 0.773 & 0.597 & 0.601\\ \hdashline
         \multicolumn{7}{l}{\textbf{Zero-shot Approaches}} \\ \hdashline
         (\textit{Spatial, Classical}) NIQE~\cite{mittal2012making} &  0.596 & 0.628 & 0.541 & 0.553 & 0.278 & 0.290 & 0.312 & 0.301\\
         (\textit{Temporal, Classical}) TPQI~\cite{tpqi} & 0.636 & 0.645 & 0.556 & 0.549 & 0.111 & 0.218 & 0.361 & 0.377\\
         (\textit{CLIP-based}) SAQI~\cite{wu2023bvqi}  & 0.629 & 0.638 & 0.608 & 0.602 & \underline{0.585} & \underline{0.606} & 0.541 & {0.559}\\ \hdashline
         (\textit{Ensemble of Three Above}) BVQI~\cite{wu2023bvqi}  & \bf{0.784} & \bf{0.794} & \underline{0.760} & \underline{0.760} & {0.525} & {0.556} & \underline{0.637} & \underline{0.648}\\ \hdashline
         \textit{Spatial}) mPO-7B (\textbf{Q-Instruct}~\cite{wu2023qinstruct}) \qquad \qquad \qquad \qquad \qquad & 0.720 & 0.657 & 0.692 & 0.603 & {0.539} & 0.540 & {0.554} & 0.515\\
         (\textit{Spatial}) mPO-7B (\textbf{Q-Boost}) & \underline{0.741} & \underline{0.793} & \bf{0.801} & \bf{0.803} & \bf{0.723} & \bf{0.705} & \bf{0.682} & \bf{0.692}\\
          \bottomrule   
    \end{tabular}}
    \label{tab:vqa}
    \vspace{-0.4cm}
\end{table*}

\subsection{Inference Cost}
Q-Boost does not introduce additional computational complexity, as the entire computation process essentially takes place after the input prompt is entered and the language backbone returns the \textit{[SCORE\_TOKEN]}. The subsequent MPE simply involves tokenizing the words used, and then calculating the \textit{logits} on the \textit{[SCORE\_TOKEN]}. As the computational cost of the softmax and score weighting of TTI is much less than the inference of the MLLM, the additional computation cost of the Q-Boost is almost negligible. This is different from CLIP-based QA methods~\cite{wang2023exploring,wu2023bvqi}, in which each prompt word needs to pass through the text backbone independently. Thus, applying a similar multi-prompt strategy to CLIP-based QA methods would significantly increase inference costs.


\section{Experiment}
\subsection{Implementation Details}
\subsubsection{Experimental Setup}
The mPLUG-Owl-2-7B~\cite{mplugowl} (\textit{abbr. as} mPO-7B) with CLIP-ViT-Large-14 vision backbone and LLaMA2-7B language backbone is used for validation. The mPO-7B weights are loaded from the instruction-tuning process of \textbf{Q-Instruct}~\cite{wu2023qinstruct} and frozen during the inference process. The mPO-7B (\textbf{Q-Instruct}) indicates using \textbf{Q-Instruct}-tuned weights without the \textbf{Q-Boost} strategy. The strategy-enhanced MLLM model is denoted as mPO-7B (\textbf{Q-Boost}). Only the TTI strategy is employed for IQA while both TTI and MPE strategies are utilized for VQA (reasons are discussed in Sec.~\ref{sec:ablation}).
In the MPE strategy, (\textit{good} + \textit{high} + \textit{fine} $\leftrightarrow$ \textit{average} + \textit{medium} + \textit{acceptable} $\leftrightarrow$ \textit{poor} + \textit{low} + \textit{bad}) are used as the groups of words for describing (\textit{positive} $\leftrightarrow$ \textit{neutral} $\leftrightarrow$ \textit{negative}) tunes. The weighted parameters $w_1$ and $w_2$ mentioned in Sec.~\ref{sec:integration}
are set as 1 and 0.5 respectively. 


\begin{figure}[t]
    \centering
    \subfigure[SRCC]{
                \centering
                \includegraphics[width = 0.8\linewidth]{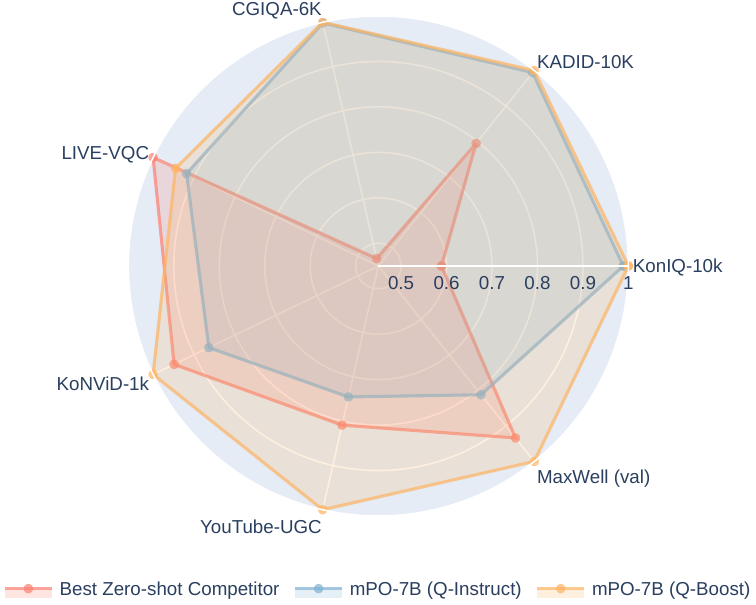}}
    \subfigure[PLCC]{
                \centering
                \includegraphics[width = 0.8\linewidth]{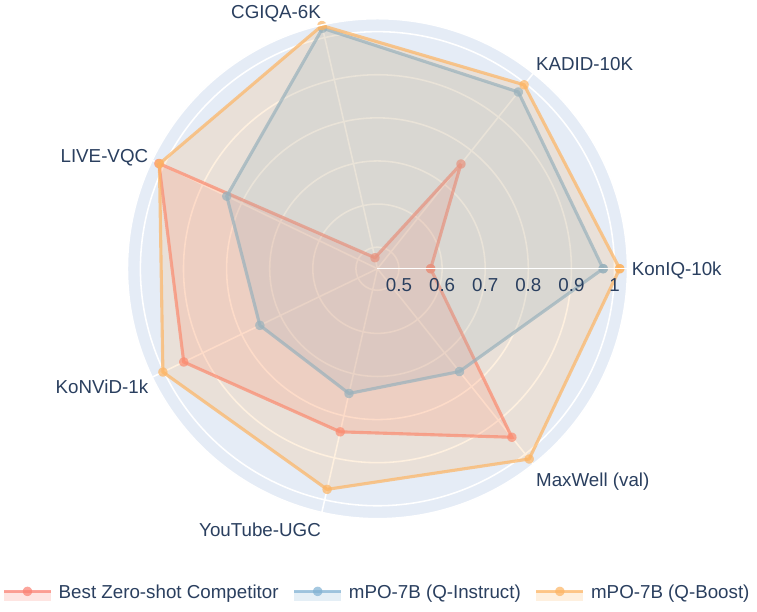}}
    \caption{The SRCC (Fig. (a)) and PLCC (Fig. (b)) performance comparison of the best zero-shot competitor, mPO-7B (\textbf{Q-Instruct}), and mPO-7B (\textbf{Q-Boost}), where the index values are calculated as $\frac{SRCC}{SRCC_{max}}$ and $\frac{PLCC}{PLCC_{max}}$. It can be seen that mPO-7B (\textbf{Q-Boost}) achieves the best performance in general and significantly boosts the performance of mPO-7B (\textbf{Q-Instruct}) on the VQA datasets.}
    \label{fig:radar}
    \vspace{-0.4cm}
\end{figure}

\label{sec:ablation}
\begin{table*}[]
    \centering
    \renewcommand\arraystretch{1.1}
    \renewcommand\tabcolsep{4pt}
    \caption{Experimental results for the ablation study. \textit{good} $\leftrightarrow$ / $\leftrightarrow$ \textit{poor} indicates using only binary extremes prompts \textit{good} and \textit{poor}, \textit{good} $\leftrightarrow$ \textit{average} $\leftrightarrow$ \textit{poor} represents using \textbf{Triadic-Tone Integration} without {\textbf{Multi-Prompt Ensemble}}, \textit{g+h+f} $\leftrightarrow$ \textit{a+m+ac} $\leftrightarrow$ \textit{p+l+b} stands for utilizing prompts of (\textit{good}+\textit{high}+\textit{fine} $\leftrightarrow$ \textit{average}+\textit{medium}+\textit{acceptable} $\leftrightarrow$ \textit{poor}+\textit{low}+\textit{bad}), which employs both strategies. }
    \resizebox{\linewidth}{!}{\begin{tabular}{l|cc:cc:cc|cc:cc:cc:cc} \toprule
         \multirow{2}{*}{\textbf{Datasets}} & \multicolumn{6}{c|}{\textbf{IQA}} & \multicolumn{8}{c}{\textbf{VQA}}\\ \cline{2-15}
         & \multicolumn{2}{c:}{\textit{KonIQ-10k}} & \multicolumn{2}{c:}{\textit{KADID-10K}} & \multicolumn{2}{c|}{\textit{CGIQA-6K}}& \multicolumn{2}{c:}{\textit{LIVE-VQC}} & \multicolumn{2}{c:}{\textit{ KoNViD-1k}} & \multicolumn{2}{c:}{\textit{YouTube-UGC}} & \multicolumn{2}{c}{\textit{MaxWell (val)}}\\ \hdashline
         \textbf{Prompts} & SRCC & PLCC & SRCC & PLCC & SRCC & PLCC & SRCC & PLCC & SRCC & PLCC & SRCC & PLCC & SRCC & PLCC\\ \hline
         \textit{good} $\leftrightarrow$ / $\leftrightarrow$ \textit{poor} & 0.902 & 0.888 & 0.698 & 0.684 & 0.629 & 0.644 & 0.720 & 0.657 & 0.692 & 0.603 & 0.539 & 0.539 & 0.554 & 0.516\\ 
         \textit{good} $\leftrightarrow$ \textit{average} $\leftrightarrow$ \textit{poor} & \bf{0.912} & \bf{0.923} & {0.702} & \bf{0.699} & \bf{0.630} & \bf{0.648} & 0.740 & 0.725 & 0.766 & 0.701 & 0.577 & 0.583 & 0.611 & 0.589\\ 
         \textit{g+h+f} $\leftrightarrow$ \textit{a+m+ac} $\leftrightarrow$ \textit{p+l+b} & {0.898} & {0.910} & \bf{0.704} & {0.678} & {0.558} & {0.577} & \bf{0.741} & \bf{0.793} & \bf{0.801} & \bf{0.803} & \bf{0.723} & \bf{0.705} & \bf{0.682} & \bf{0.692}\\ 
          \bottomrule   
    \end{tabular}}
    \label{tab:ablation}
    \vspace{-0.4cm}
\end{table*}

\subsubsection{Datasets Introduction}
3 IQA datasets and 4 VQA datasets are utilized for performance comparison, which are briefly reviewed in Table~\ref{tab:datasets}. About 48.92\% images of the KonIQ-10k~\cite{koniq} dataset are included in the \textbf{Q-Instruct}~\cite{wu2023qinstruct}-tuning stage and no images of the rest datasets are seen by the mPO-7B (\textbf{Q-Instruct} \& \textbf{Q-Boost}). Additionally, the CGIQA-6K ~\cite{zhang2023subjective} dataset contains two separate sub-sets which consist of 3,000 game images and 3,000 movie images respectively and we average the performance of the sub-sets for the final exhibition.

\subsubsection{IQA/VQA Competitors}
Both opinion-aware (need supervised training) and zero-shot methods (training-free) are included for comparison. \\
$\bullet$ The opinion-aware IQA methods consist of DBCNN~\cite{dbcnn}, HyperIQA \cite{su2020blindly}, and MUSIQ \cite{ke2021musiq}. \\
$\bullet$ The zero-shot IQA methods contain NIQE~\cite{mittal2012making}, IL-NIQE~\cite{zhang2015feature}, and CLIP-IQA~\cite{wang2023exploring}. \\
$\bullet$ The opinion-aware VQA methods include TLVQM~\cite{tlvqm}, VSFA~\cite{vsfa}, and VIDEVAL~\cite{videval}.\\
$\bullet$ The zero-shot VQA methods contain NIQE~\cite{zhang2015feature}, TPQI~\cite{tpqi}, and SAQI~\cite{wu2023bvqi}.

\subsection{Performance Discussion}
The experimental performance on the IQA and VQA datasets is illustrated in Table~\ref{tab:iqa} and Table~\ref{tab:vqa}, from which we can obtain several interesting findings. \\
\textbf{Finding 1}: Both mPO-7B (\textbf{Q-Instruct}) and mPO-7B (\textbf{Q-Boost}) have outperformed the existing zero-shot NR-IQA methods by a large margin and are even competitive with the option-aware approaches on the IQA datasets. This reveals the powerful potential of MLLM in IQA tasks. Moreover, compared to traditional MOS (mean opinion score) regression training models, MLLM is text-supervised, which means that using text supervision instead of MOS regression can achieve similar or even better zero-shot performance. This also opens up a possible new training path in the QA field, allowing us to train and enhance AI's QA capabilities in a more intuitive and explanatory manner. \\
\textbf{Finding 2}: The mPO-7B (\textbf{Q-Boost}) significantly advances the mPO-7B (\textbf{Q-Instruct}) and achieves impressive zero-shot performance on the VQA datasets even without considering temporal quality information. The mPO-7B (\textbf{Q-Boost}) model outperforms all compared methodologies, encompassing supervised VQA techniques, on both the KoNViD-1k and MaxWell (val) datasets. This demonstrates the superior assessment capabilities and remarkable generalization proficiency of MLLM when augmented with \textbf{Q-Boost} in the realm of VQA. The progress of \textbf{Q-Boost} stems from leveraging the advantages of diversity, eliminating biases, and providing a more comprehensive and thorough assessment.

\subsection{Ablation Study}
To quantify the contributions of the TTI and MPE strategies, we conduct the ablation study in this section, the results of which are shown in Table~\ref{tab:ablation}. 1) It is observed that the TTI strategy yields a modest enhancement in the performance of IQA), whereas the MPE strategy appears to have negligible impact. This phenomenon could be attributed to the more explicit nature of IQA, wherein the ensemble of prompts may inadvertently introduce elements of confusion. Consequently, we recommend the exclusive use of TTI as the \textbf{Q-Boost} approach for optimizing IQA tasks. 2) Both strategies make significant contributions to enhancing the VQA understanding of MLLM. This is because the MLLM's weights are pre-trained with image-text pairs, with a gap to the video data. Thus these strategies effectively mitigate biases arising from this data gap and utilize the diversity of prompts to enable more accurate and comprehensive assessment.

\section{Conclusion}
In conclusion, this paper has made significant strides in bridging the gap in the application of MLLM for visual quality assessment, a crucial area in the realm of low-level vision tasks. Our introduction of \textbf{Q-Boost} marks a pivotal advancement in this field. The unique incorporation of \textbf{Triadic-Tone Integration}, which departs from the traditional binary prompt approach to include a neutral perspective, provides a more nuanced and balanced evaluation. Additionally, the \textbf{Multi-Prompt Ensemble} technique effectively mitigates biases and enhances assessment accuracy. The experimental results unequivocally illustrate that MLLM equipped with the \textbf{Q-Boost} strategy exhibits outstanding zero-shot performance on both IQA and VQA tasks.  We believe that the \textbf{Q-Boost} strategy will serve as a catalyst for further advancements in the quality assessment field and inspire new methodologies in the broader context of MLLM for low-level vision.
\bibliographystyle{IEEEbib}
\bibliography{icme2023template}

\end{document}